\def\year{2022}\relax
\documentclass[letterpaper]{article} 
\usepackage{aaai22}  
\usepackage{times}  
\usepackage{helvet}  
\usepackage{courier}  
\usepackage[hyphens]{url}  
\usepackage{graphicx} 
\usepackage{natbib}  
\usepackage{caption} 
\DeclareCaptionStyle{ruled}{labelfont=normalfont,labelsep=colon,strut=off} 
\usepackage{amsmath}
\usepackage{amssymb}
\usepackage{multirow}
\usepackage{booktabs}
\usepackage{bbding}

\title{Domain Adaptation on Semantic Segmentation with Separate Affine Transformation in Batch Normalization}
\author {
    Junhao Yan,\textsuperscript{\rm 1}
    Wonsook Lee \textsuperscript{\rm 2}
}
\affiliations {
    \textsuperscript{\rm 1} University of Ottawa\\
    \textsuperscript{\rm 2} University of Ottawa\\
    jyan033@uottawa.ca, wslee@uottawa.ca
}

\begin{document}
\maketitle
\begin{abstract}
In recent years, unsupervised domain adaptation (UDA) for semantic segmentation has brought many researchers’ attention. Many of them take an approach to design a complex system so as to better align the gap between source and target domain. Instead, we focus on the very basic structure of the deep neural network, Batch Normalization, and propose to replace the Sharing Affine Transformation with our proposed Separate Affine Transformation (SEAT). The proposed SEAT is simple, easily implemented and easy to integrate into existing adversarial learning based UDA methods. Also, to further improve the adaptation quality, we introduce multi level adaptation by adding the lower-level features to the higher-level ones before feeding them to the discriminator, without adding extra discriminator like others. Experiments show that the proposed methods is less complex without losing performance accuracy when compared with other UDA methods. 
\end{abstract}
\begin{figure*}[ht]
    \centering
    
    \includegraphics[width=0.7\textwidth]{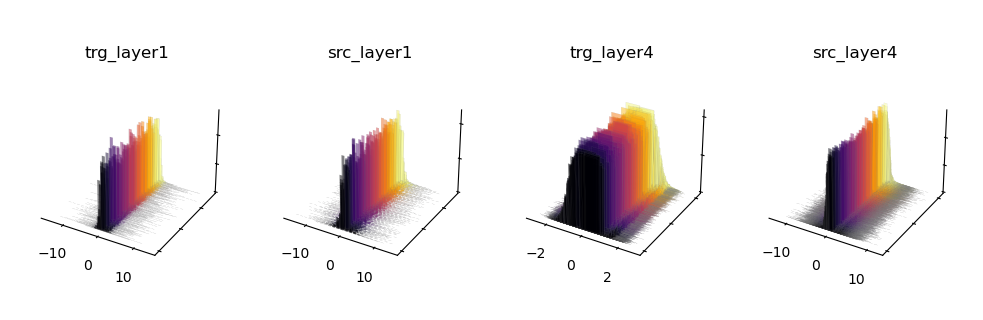}
    \caption{The normalized source domain and target domain feature distribution in different layers. The experiment is based on ResNet101.}
    \label{fig:distribution}
\end{figure*}
\section{Introduction}
Annotated large datasets \cite{lin2014microsoft, deng2009imagenet} have enabled great progress in semantic segmentation \cite{long2015fully} in Deep Learning. Unfortunately, collecting and manually annotating large datasets with dense pixel level labels has been extremely costly due to the large amount of human effort required. Moreover, manual annotation always comes with label noise like an average of 3.4\% label errors at testing set across 10 commonly used datasets \cite{northcutt2021pervasive}. An appealing idea is to utilize synthetic data that is automatically generated and labeled by the computer. However, because the distribution between source, with synthetic data, and target, with  real world data, domains are different \cite{hoffman2016fcns}, which results in poor performance in target domain data when trained with source domain data. \\
To address this issue, the UDA technique was firstly proposed by Ben-David et al \cite{ben2007analysis} to reduce domain shift problem. Over time, UDA has also been utilized on the semantic segmentation task and brought researchers attentions. In order to extract domain invariant features, some of them utilize preprocessed source data as inputs \cite{hoffman2018cycada, Yang_2020_CVPR} and different textures source data as the network's input \cite{kim2020learning}, etc, while others choose to add more constraints to the network, like only similar objects between source images and target images are aligned \cite{wang2020differential, kang2020pixel}, etc.\\
It is common to share BatchNorm (BN) layers \cite{ioffe2015batch} even though source domain and target domain have different distributions. This problem was partially resolved by modulating the statistics from the source domain to the target domain in all BN layers across the network \cite{li2018adaptive}. Another approach \cite{2019transferable} proposes TransNorm, where the mean and variance of each channel of inputs are computed separately and the learnable domain sharing affine transformations are used to scale and shift the normalized values. \\
We observe the distribution of source and target domain features over a batch after normalization but before affine transformation step inside one BN layer during training, which is shown at figure \ref{fig:distribution}. We find that the distribution between them is not exactly the same, similar looking though. Thus, it is better to use separate affine transformations for different distributions. So we propose to use separate normalizing and affine transformation (SEAT) at batch normalization layers for source domain inputs and target domain inputs respectively so as to solve the aforementioned problem.\\
As mentioned in \cite{tsai2018learning}, lower-level features are not adapted well. Different from directly adding one more discriminator to adapt lower-level features, we propose to sum lower-level features with higher-level features with a hyper-parameter, which saves more computational resource during training comparing with directly adding one more discriminator like others.\\
Our innovation comes in SEAT and multi level adaptation which are useful and require less complexities, but still shows comparable domain adaption results with others. Self training and transferred style source images are also adapted in our experiment to further improve model’s performance. 
\section{Related Works}
\textbf{Domain Adaptation:} Adversarial learning is widely used on UDA area \cite{tsai2018learning, hoffman2016fcns, hoffman2018cycada, zhang2017curriculum, ganin2016domain, wang2020differential}. Unlike the classification task that adapts on the feature space \cite{long2015learning, ganin2016domain}, \citet{tsai2018learning} proposes to do adaptation on structured output space for the semantic segmentation task because there are less visual cues and details that needed to adapt comparing with the classification task. Based on previous work from \cite{tsai2018learning}, more and more researches start to make innovations in this area. For example, \citet{hoffman2018cycada} utilize transferred target style source domain images generated from CycleGan \cite{Zhu_2017_ICCV} to train the model. \citet{kim2020learning} diversifies the texture of source domain images and then the generated images with various textures are used during training so as to prevent the segmentation model from overfitting to on specific texture. In order to better align cross domain features, more constraints are also added during training. For example, \citet{vu2019advent} combines entropy based loss with adversarial loss to better align the domain gap, while \citet{pan2020unsupervised}, based on the original adversarial training based UDA methods, proposes to split the target domain images into hard and easy split using an entropy-based ranking function. And then fine tune the model by adapting the domain from easy split to hard split. In order to adapt the target domain features towards the most likely space of source domain features, \citet{wang2020differential} proposes to treat stuff categories (sky, tree, etc) and thing categories (car, traffic sign, etc) with different strategies.  \citet{li2020content} facilitates Content-Consistent Matching so as to find out synthetic images with similar distribution as real ones in target domain. \\
\textbf{Self Training:} Because of the power of self training, many UDA methods \cite{Yang_2020_CVPR, wang2020differential, kim2020learning} adopt self training as their second training stage to further improve model performance. As UDA methods lack GT labels for target domain images, which results in poorer performance than supervised ones. And what self training does is to retrain the whole model with pseudo labels on target domain generated by previous model's confidence prediction. In order to get comparative results, we also adopt self training as second stage training. 
\section{Methods}
\begin{figure*}[ht]
    \centering
    
    \includegraphics[width=0.7\textwidth]{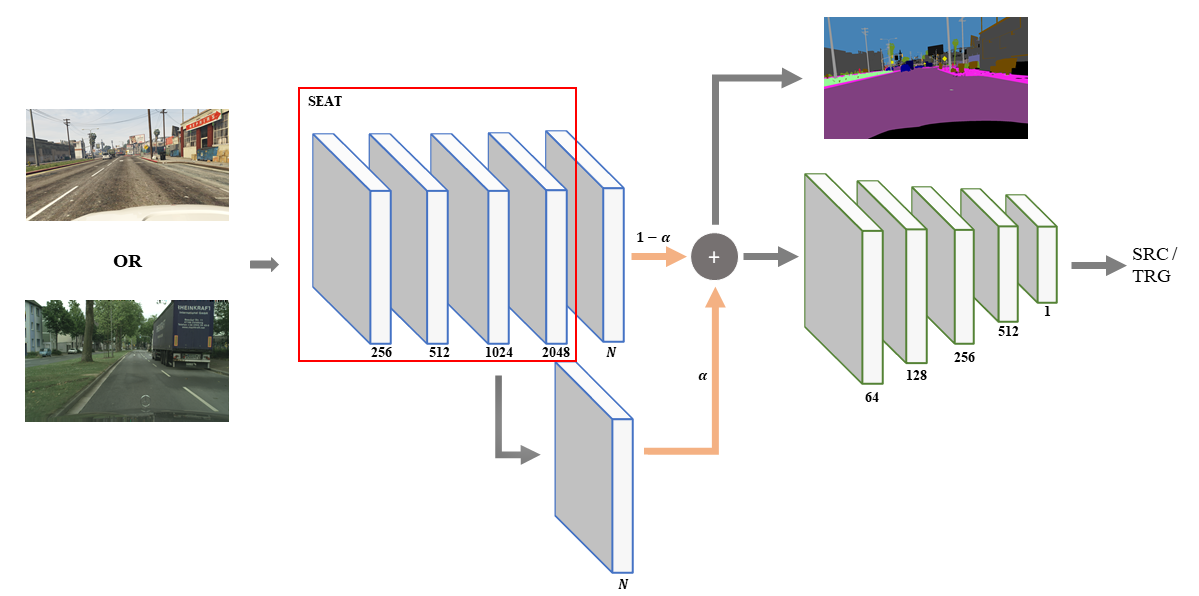}
    \caption{The overall model architecture. Blue blocks represent $\mathbf{G}$ integrating with our proposed SEAT inside the read rectangle. Green blocks represent $\mathbf{D}$. Arrows correspond the flow direction, where orange ones represent our multi level adaptation by adding the lower-level features with higher-level ones with parameter $\alpha$. Numbers represent size of channels, while $\mathcal{N}$, $\oplus$ denotes the number of classes and the sum operation.}
    \label{fig:model archit}
\end{figure*}

Our proposed method consists of two model architectures, which are the segmentation network $\mathbf{G}$ integrating with SEAT and the discriminator $\mathbf{D}$ respectively. The overall model architecture is shown at figure \ref{fig:model archit}.\\
\subsection{Objective Functions}
The source domain images $x^s\in\mathbb{R}^{\mathcal{\text{3}\times{H}\times{W}}}$ and target domain images $x^t\in\mathbb{R}^{\mathcal{\text{3}\times{H}\times{W}}}$ are firstly forwarded to the $\mathbf{G}$ to get lower-level and higher-level features of source domain and target domain respectively, denoted as \{$f^{ls}$, $f^{hs}$, $f^{lt}$, $f^{ht}$\}$\in\mathbb{R}^\mathcal{\text{19}\times{H}\times{W}}$. With those features, the following loss functions are used to train $\mathbf{D}$ and $\mathbf{G}$. \\
\textbf{Discriminator Objective Function:} $\mathbf{D}$ is trained to distinguish source and target features. Thus, the binary cross-entropy loss is applied and shown at equation \ref{eq:dis}.
\begin{center}
    \begin{align}
        \mathcal{L}_{dis}  = -\frac{1}{\mathcal{M}}\sum_{i=0}^\mathcal{M}&\mathcal{L}^{src}_{dis}(\alpha{f^{ls}_i}+(1-\alpha)f^{hs}_i)\nonumber\\ &+ \mathcal{L}^{trg}_{dis}(\alpha{f^{lt}_i}+(1-\alpha)f^{ht}_i) \label{eq:dis}
    \end{align}
\end{center}
where
\begin{center}
    \begin{align}
        \mathcal{L}^{src}_{dis}(x) &=
        (1-y)log(1-\mathbf{D}(x))  \nonumber\\
        \mathcal{L}^{trg}_{dis}(x) &=
        ylog(\mathbf{D}(x))\nonumber
    \end{align}
\end{center}
$y=1$ if samples are drawn from target domain, $y=0$ if samples are drawn from source domain. $\mathcal{M}$ represents the number of images in a batch. \\
\textbf{Segmentation Network Objective Functions:} $\mathbf{G}$ needs to do two things right here. First, we apply adversarial loss such that $\mathbf{G}$ is able to output features in source style when $x^t$ are given, which is shown at the equation \ref{eq:eqgen}. 
\begin{center}
    \begin{align}
        \mathcal{L}_{adv}=-\frac{1}{\mathcal{M}}\sum_{i=0}^\mathcal{M}log(1-\mathbf{D}(\alpha{f^{lt}_i}+(1-\alpha)f^{ht}_i)) \label{eq:eqgen}
    \end{align}
\end{center}
Second, given $x^s$, the cross-entropy loss is also applied to $\mathbf{G}$ so as to predict segmentation map, which is shown at the equation \ref{eq:seg}.
\begin{center}
    \begin{align}
        \mathcal{L}_{seg}=-\frac{1}{\mathcal{M}}\sum_{i=0}^{\mathcal{M}}Y^s_ilog(\alpha{f^{ls}_i}+(1-\alpha)f^{hs}_i)\label{eq:seg}
    \end{align}
\end{center}
where $Y^s$ represents source GT labels.\\
The overall loss function of $\mathbf{G}$ is shown at equation \ref{eq:ssn}.
\begin{center}
    \begin{align}
        \mathcal{L}_{ssn} = \beta\mathcal{L}_{adv} + \mathcal{L}_{seg}\label{eq:ssn}
    \end{align}
\end{center}
where $\beta$ is set to 0.001 during experiments.
\subsection{Separate Affine Transformation}
\begin{figure}[ht]
    \centering
    \includegraphics[width=0.7\columnwidth]{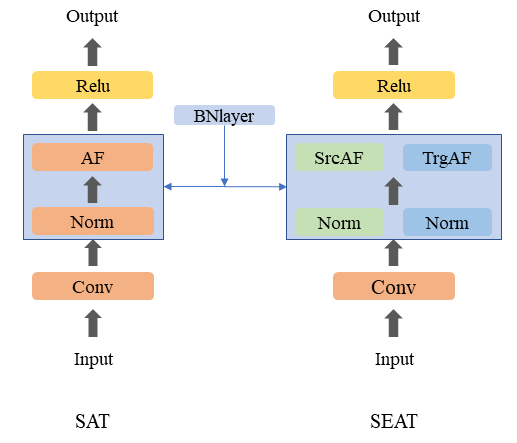}
    \caption{Difference between SAT (Conventional one) and SEAT (Proposed one).}
    \label{fig: separate bn}
\end{figure}

In this section, we will describe why Sharing Affine Transformation (SAT) degrades the model performance and how SEAT avoids this issue.\\
First, we want to build up the relation between the Cross-Entropy and the KL-Divergence \cite{cover1999elements}, where KL-Divergence is used to measure the distance between two probability distributions. The definitions for them are shown below.
\begin{center}
    \begin{align}
        H(A,B)&=-\sum_xP_A(x)log(P_B(x))\label{eq: ce}\\
        KL(A,B)&=\sum_xP(A)(x)log(P_A(x))\nonumber\\
        &-P_A(x)log(P_B(x))\label{eq: kl}
    \end{align}
\end{center}
where $A$ and $B$ represent GT's distribution and prediction's distribution here.\\
If we substitute equation \ref{eq: ce} with equation \ref{eq: kl}, we get equation \ref{eq: cekl}.
\begin{center}
\begin{align}
    H(A,B)=KL(A,B)+H(A,A)\label{eq: cekl}
\end{align}
\end{center}
From equation \ref{eq: cekl}, as $H(A,A)$ is a constant. Thus, we could consider that minimizing the Cross-Entropy is equal to minimizing the distance between probability distribution $A$ and $B$.
\begin{figure*}[ht]
\centering
\begin{tabular}{ccc}
    \includegraphics[width=0.2\textwidth]{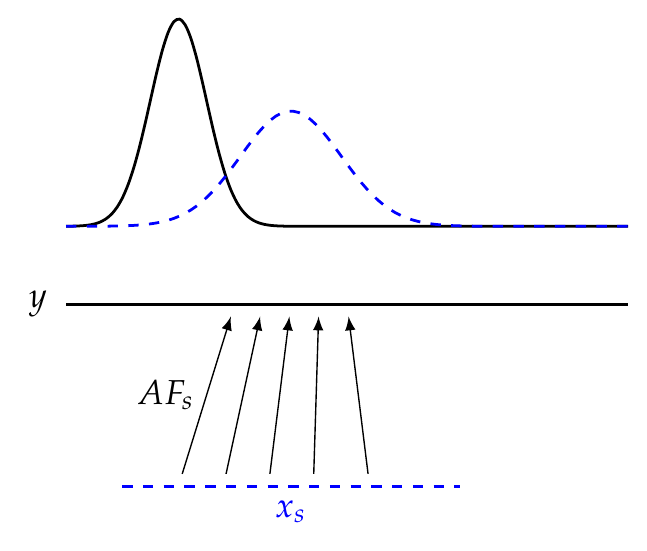} 
    &  
    \includegraphics[width=0.2\textwidth]{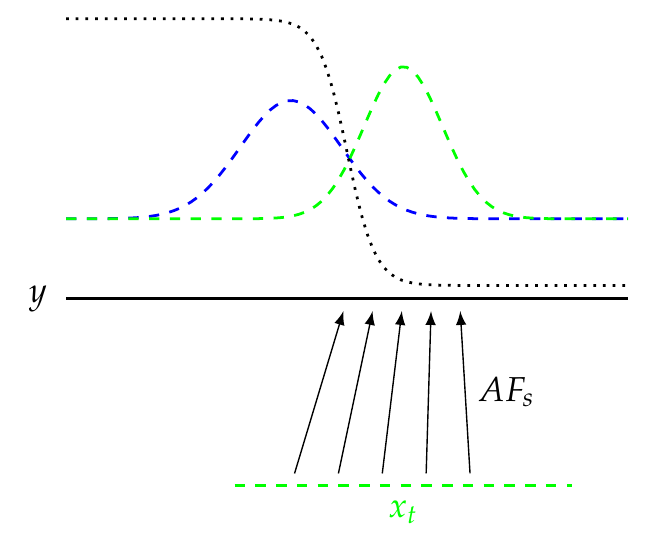}
    & 
    \includegraphics[width=0.2\textwidth]{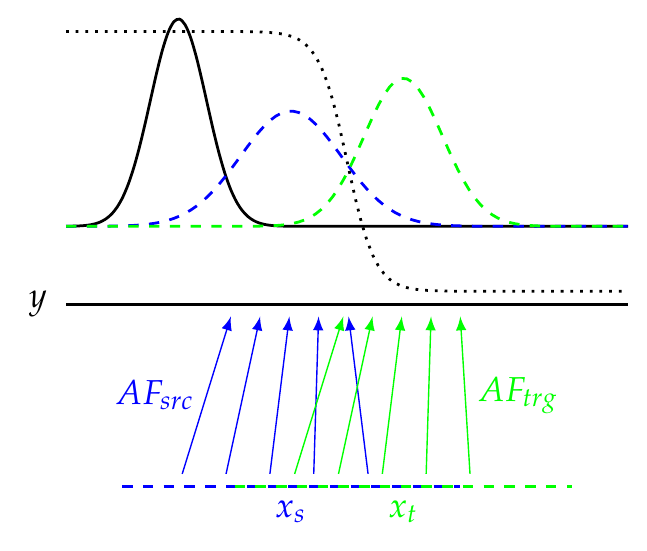}
    \\
    \centering (a)
    & 
    \centering (b) 
    & 
    \centering (c) 
\end{tabular}
\caption{The upward arrows represent Affine Transformation inside BN layers, which maps normalized input $x$ to output $y$ ($y=AF(x)$). SAT is trained simultaneously to align distribution between (a) Source GT (black, solid line) and source's prediction (blue, dashed line) under the guide of the Cross-Entropy loss (proved at equation \ref{eq: cekl}) given source domain input (blue, horizontal, dashed line), and to align the distribution between (b) Source's prediction and target's prediction (green, dashed line) guided by discriminative distribution (black, dotted line) given target domain input (green, horizontal, dashed line). (c) SEAT uses separate affine transformations for source and target domain input.}
\label{fig: jatseat_dis}
\end{figure*}\\
Let's go back to the issue brought from the SAT. In order to give a straightforward explanation, we draw how affine transformation maps input distribution to output distribution during training at figure \ref{fig: jatseat_dis}. As we could see from the figure \ref{fig: jatseat_dis} (a) and (b), the update of SAT is conflicting if SAT is trained simultaneously to align distribution between source's prediction to that of source GT and to align the distribution of target's prediction to that of source's prediction, which results in negative impact on the domain adaptation. \\
On the contrary, if we utilize source and target Affine Transformation ($AF_{src}$ and $AF_{trg}$) for source domain inputs and target domain inputs shown at the figure \ref{fig: jatseat_dis} (c), the conflict is avoided. The difference between BN layers with SAT and SEAT is shown at figure \ref{fig: separate bn}.
\subsection{Self-Training}
After the first stage training introduced above, we could use the trained model to generate pseudo labels for target domain images given a certain threshold $\psi\in[0,1]$. More specifically, $\mathbf{G}$ is initialized and trained with equation \ref{eq:ssn} at stage 1 without any target domain's labels involved. After that, for each image in target domain train set, the trained $\mathbf{G}$ from stage 1 is used to predict semantic segmentation map $\phi\in\mathbb{R}^{\mathcal{N}\times\mathcal{H}\times\mathcal{W}}$. With $\phi$, we could generate pseudo label $Y^t\in\mathbb{R}^{\mathcal{H\times{W}}}$ by the following equation.
\begin{equation}
\centering
Y^t[i,j]=\left\{
             \begin{array}{lr}
             \kappa,\,if\, \phi[\kappa,i,j]\geq\psi&  \\
             255\\
             \end{array}
\right.
\end{equation}
where i$\in[0, \mathcal{H}]$, j$\in[0,\mathcal{W}]$ and 
\begin{equation}
    \centering
    \kappa=argmax(\phi[:,i,j]) \nonumber
\end{equation}
Given pseudo labels $Y^t$, we could reinitialize and train $\mathbf{G}$ at stage 2 by equation \ref{eq:sst}.
\begin{equation}
    \centering
        \mathcal{L}_{ssn} = \beta\mathcal{L}_{adv} + \mathcal{L}_{seg} + \mathcal{L}_{st}\label{eq:sst}
\end{equation}
where
\begin{equation}
    \centering
    \mathcal{L}_{st}=-\frac{1}{\mathcal{M}}\sum_{i=0}^{\mathcal{M}}Y^t_ilog(\alpha{f^{lt}_i}+(1-\alpha)f^{ht}_i)\nonumber
\end{equation}
\section{Implementation}
\subsection{Datasets}
We evaluate our method on three commonly used datasets in this area, which are GTA5 \cite{richter2016playing}, Synthia \cite{ros2016Synthia} and Cityscapes \cite{cordts2016cityscapes} respectively, where the GTA5 and Synthia are source domain dataset with semantic labels, while the Cityscapes is the target domain dataset without semantic labels. The two adaptation scenarios are GTA5 $\to$ Cityscapes and Synthia $\to$ Cityscapes respectively.\\
\textbf{GTA5:} It contains pixel-accurate semantic label maps for images, which is created from a famous computer game named GTA5. It contains 24,966 synthetic images with 33 classes of semantic annotations. During training, 19 classes in common are used and input images are resized into [1280, 720] and then randomly crop into [1024, 512].\\
\textbf{Synthia:} We choose to use the Synthia-RAND-CITYSCAPES subset, which contains 9,400 images. During training, we resize the images into [1280, 760] and then randomly crop the images into [1024, 512]. For Synthia dataset, 16 common classes are used to train, while 13 common classes are used on evaluation for standard comparison.\\
\textbf{Cityscapes:} A real world dataset, which is collected from driving scenarios, consists of 30 classes but 19 classes are used only during evaluation for standard comparison. We use 2,975 images to train our model and 500 images for evaluation. In order to keep consistent with synthetic datasets, we resize the input image into [1024, 512].\\
\subsection{Models}
DeepLabV2 \cite{chen2017deeplab} with ResNet101 \cite{he2016deep} as backbone is used as the segmentation network, while other papers also use VGG16 \cite{simonyan2014very} as backbone to verify their methods. However, the version they use contains no BN layers, it is time-consuming to re-implement and make comparison. Thus, ResNet101 is used as model backbone only in this paper. As for discriminator, we use same model structure as \cite{tsai2018learning}.\\
\subsection{Training Details}
We train our models on a RTX3090 with batch size of 1. For the segmentation network, the optimizer we use is SGD with initial learning rate of 2.5e-4, weight decay of 5.0e-4 and momentum of 0.9. As for the discriminator, we use Adam with initial learning rate of 1e-4 and the betas is set to (0.9, 0.99). The polynomial learning rate scheduler with power of 0.9 is applied to all models to adjust learning rate.
\begin{figure*}[htbp]

\begin{tabular}{cccc}
    \includegraphics[width=0.22\textwidth]{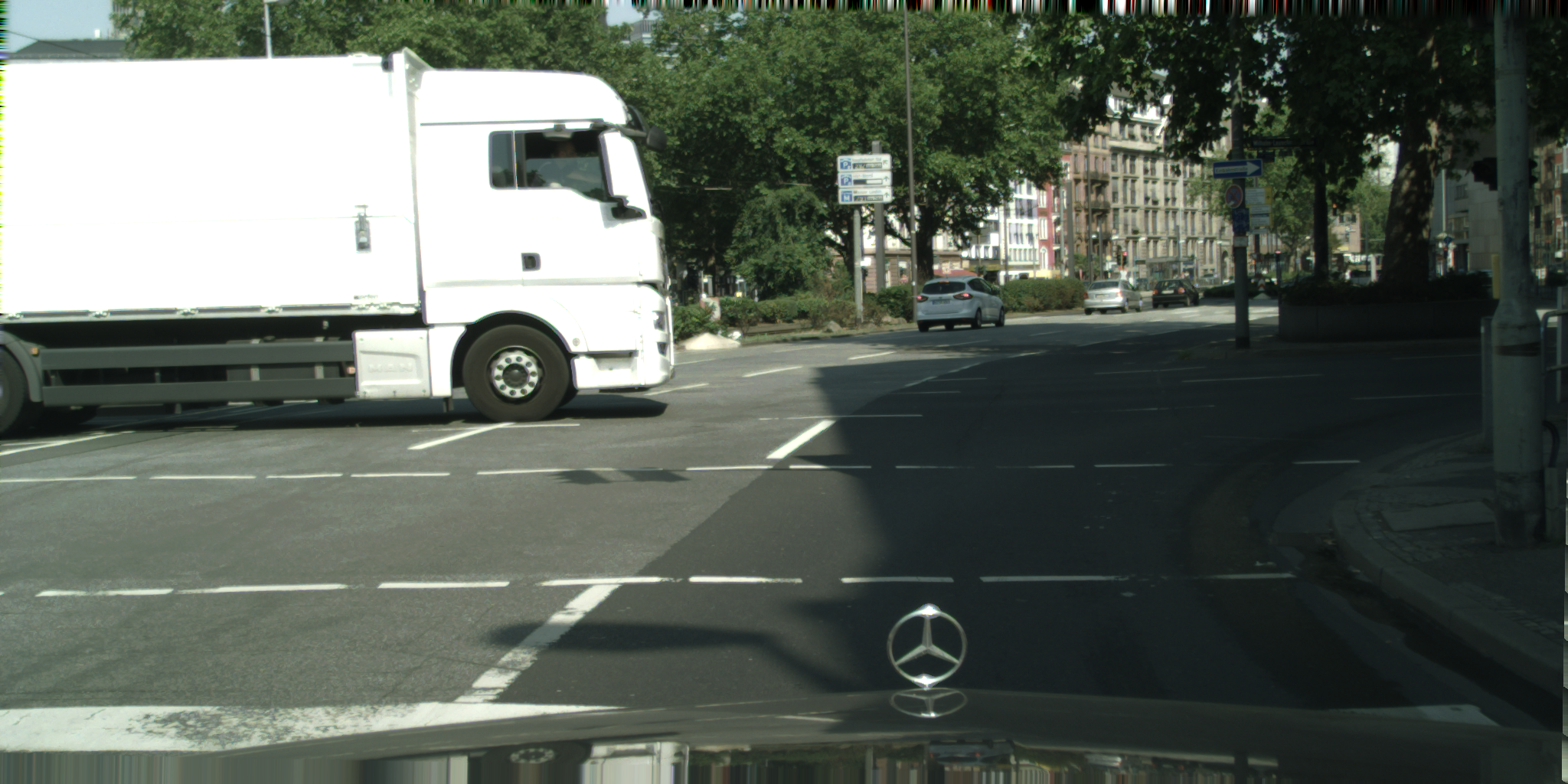} 
    &  
    \includegraphics[width=0.22\textwidth]{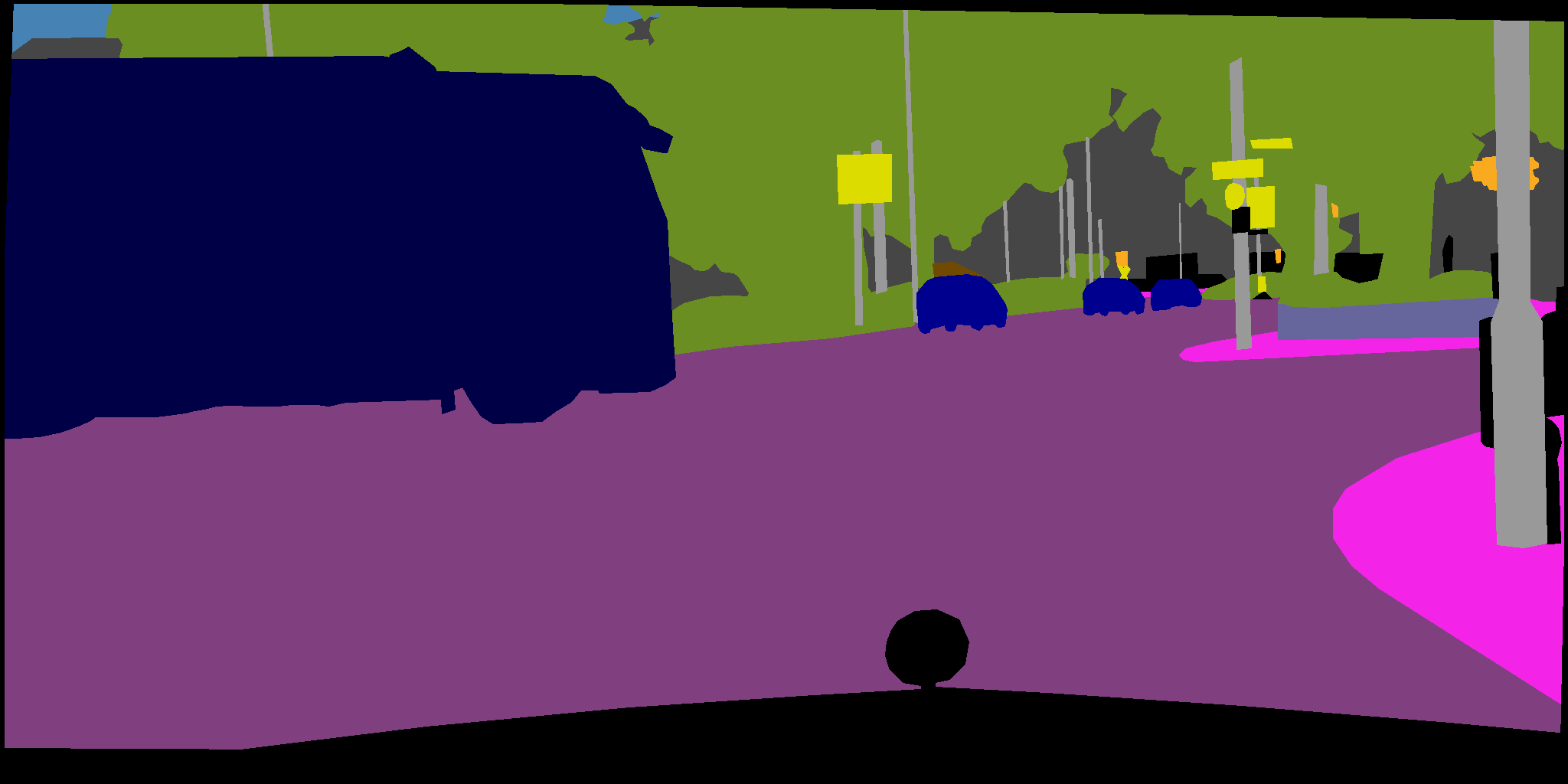}
    & 
    \includegraphics[width=0.22\textwidth]{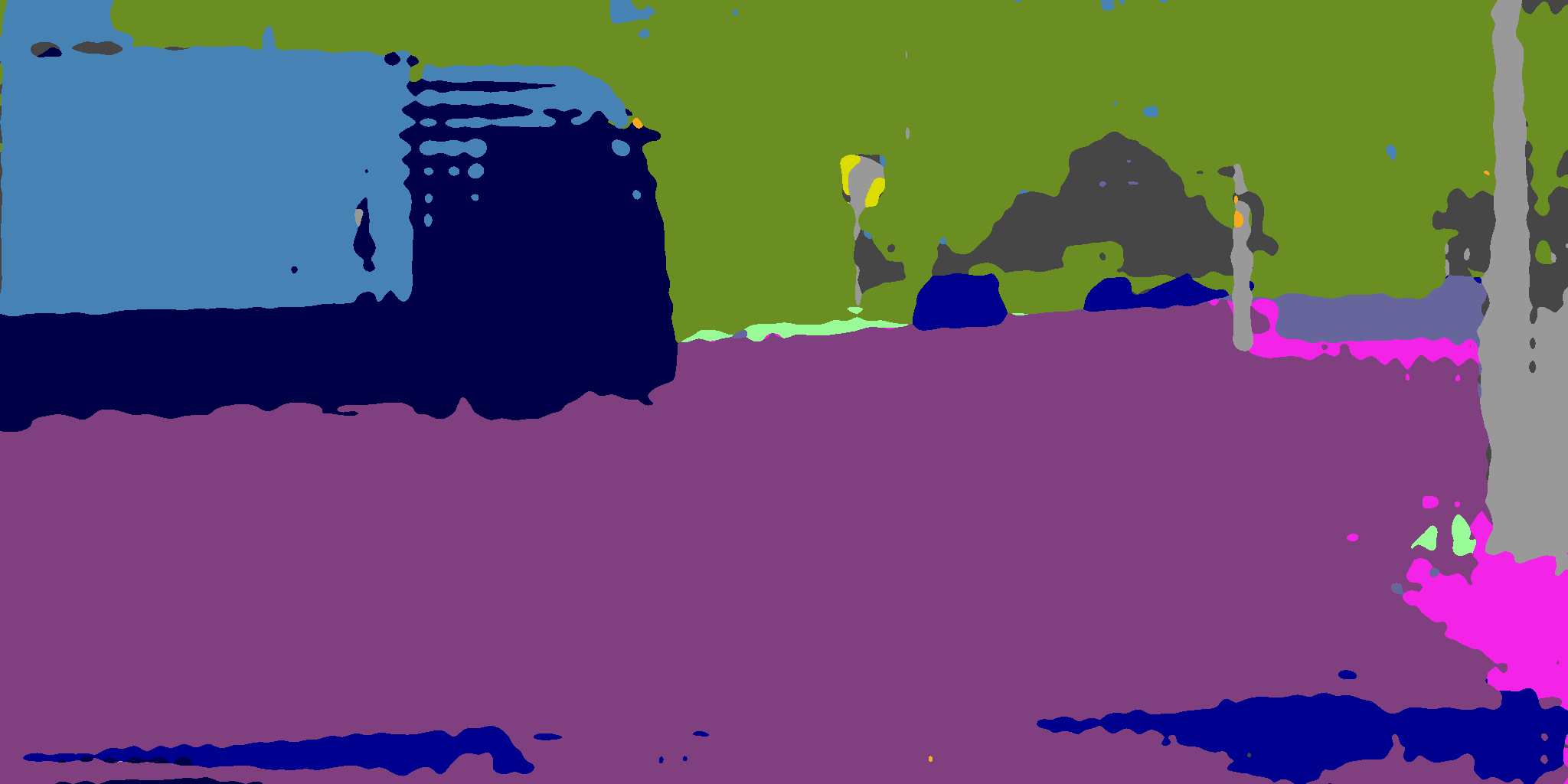}
    &
    \includegraphics[width=0.22\textwidth]{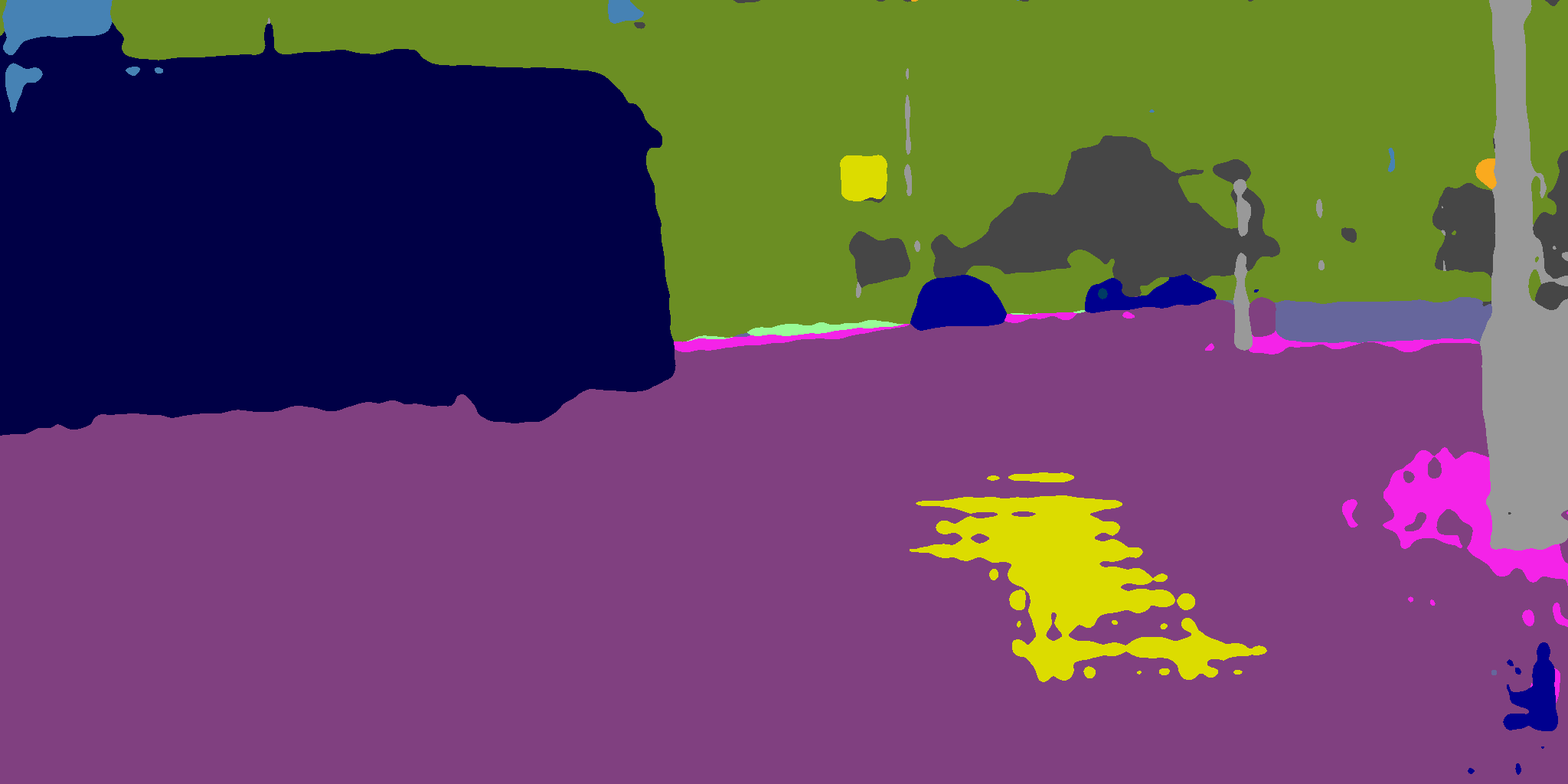}
    \\
    \includegraphics[width=0.22\textwidth]{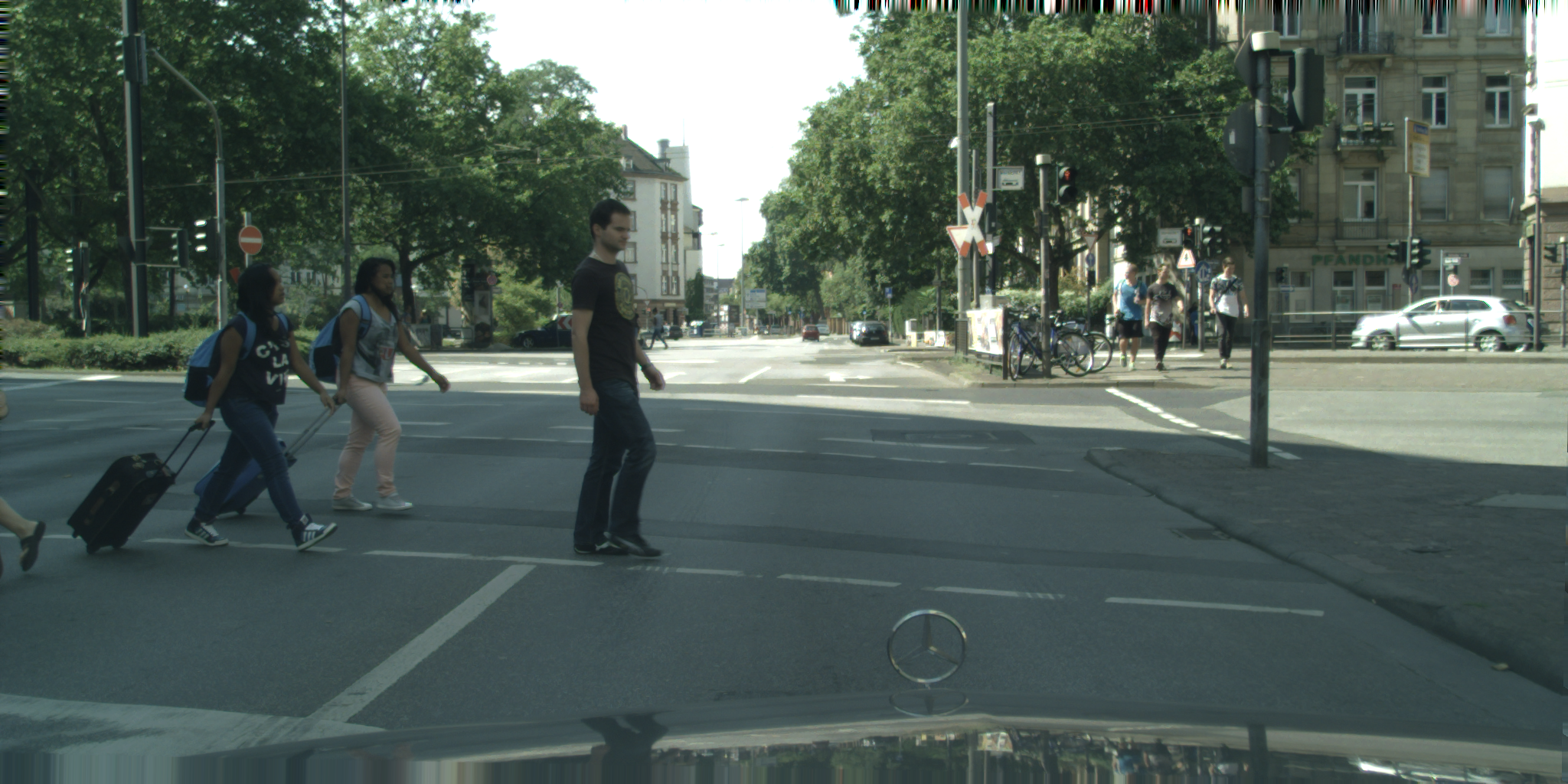} 
    &  
    \includegraphics[width=0.22\textwidth]{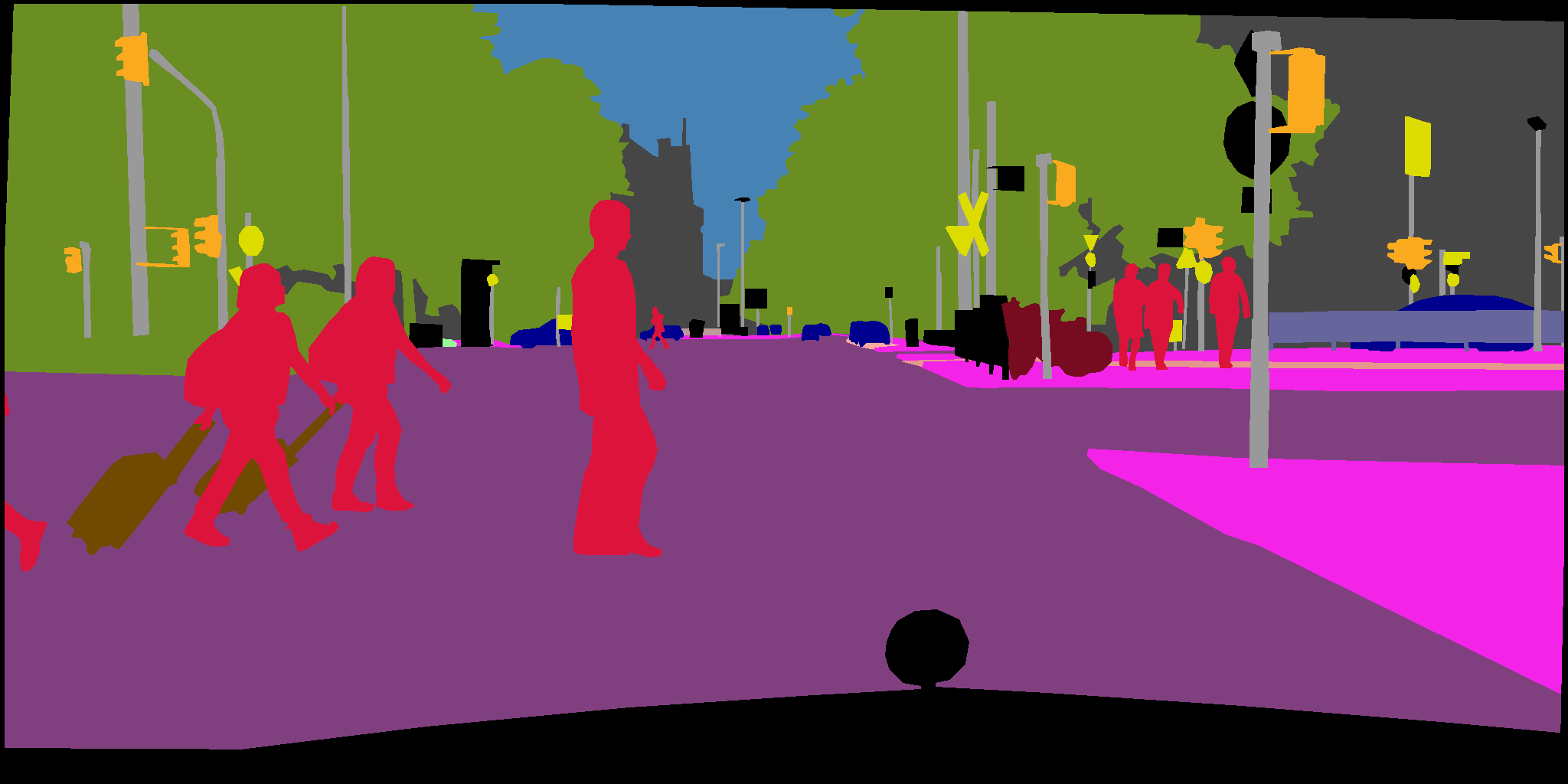}
    & 
    \includegraphics[width=0.22\textwidth]{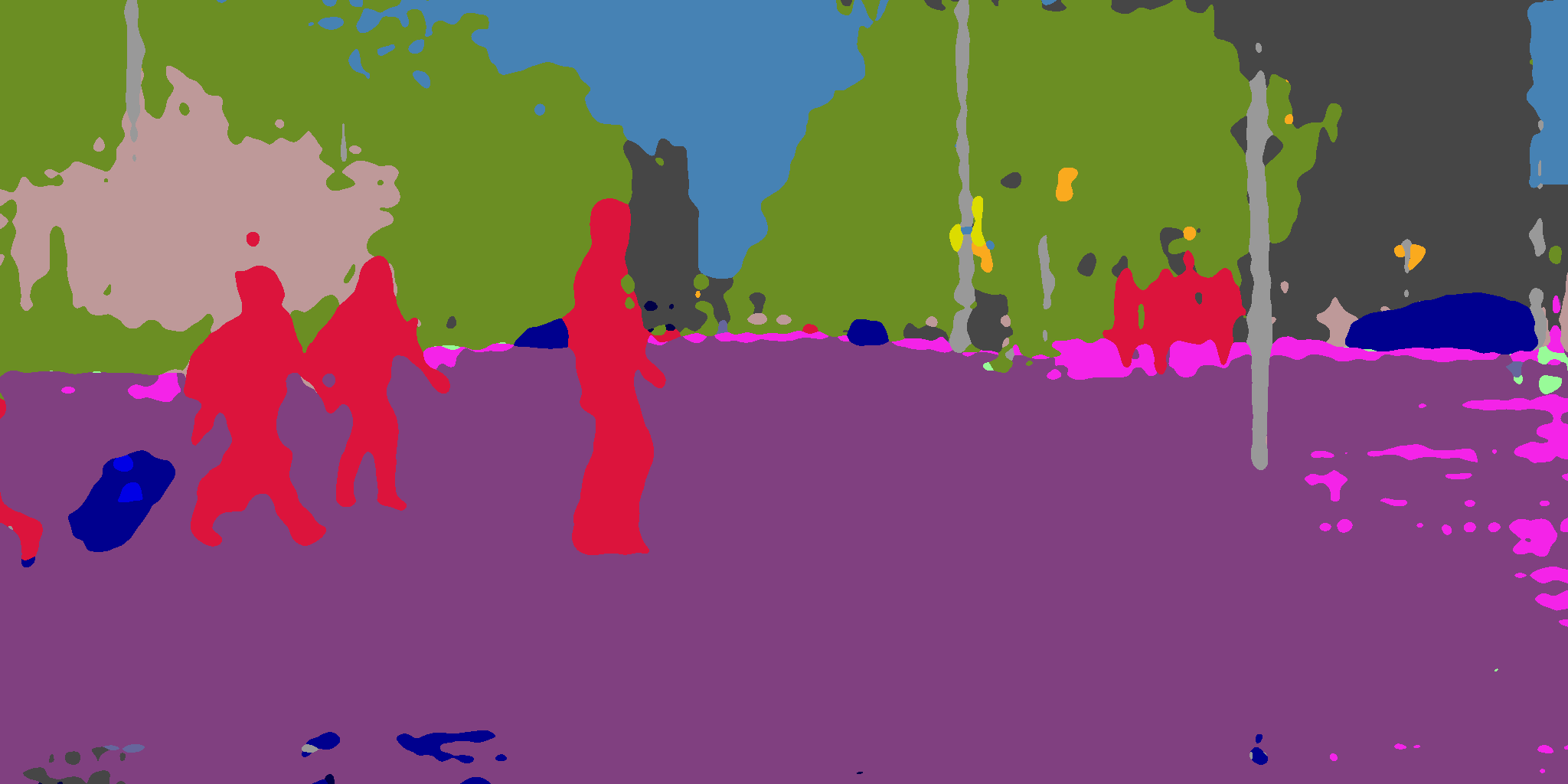}
    &
    \includegraphics[width=0.22\textwidth]{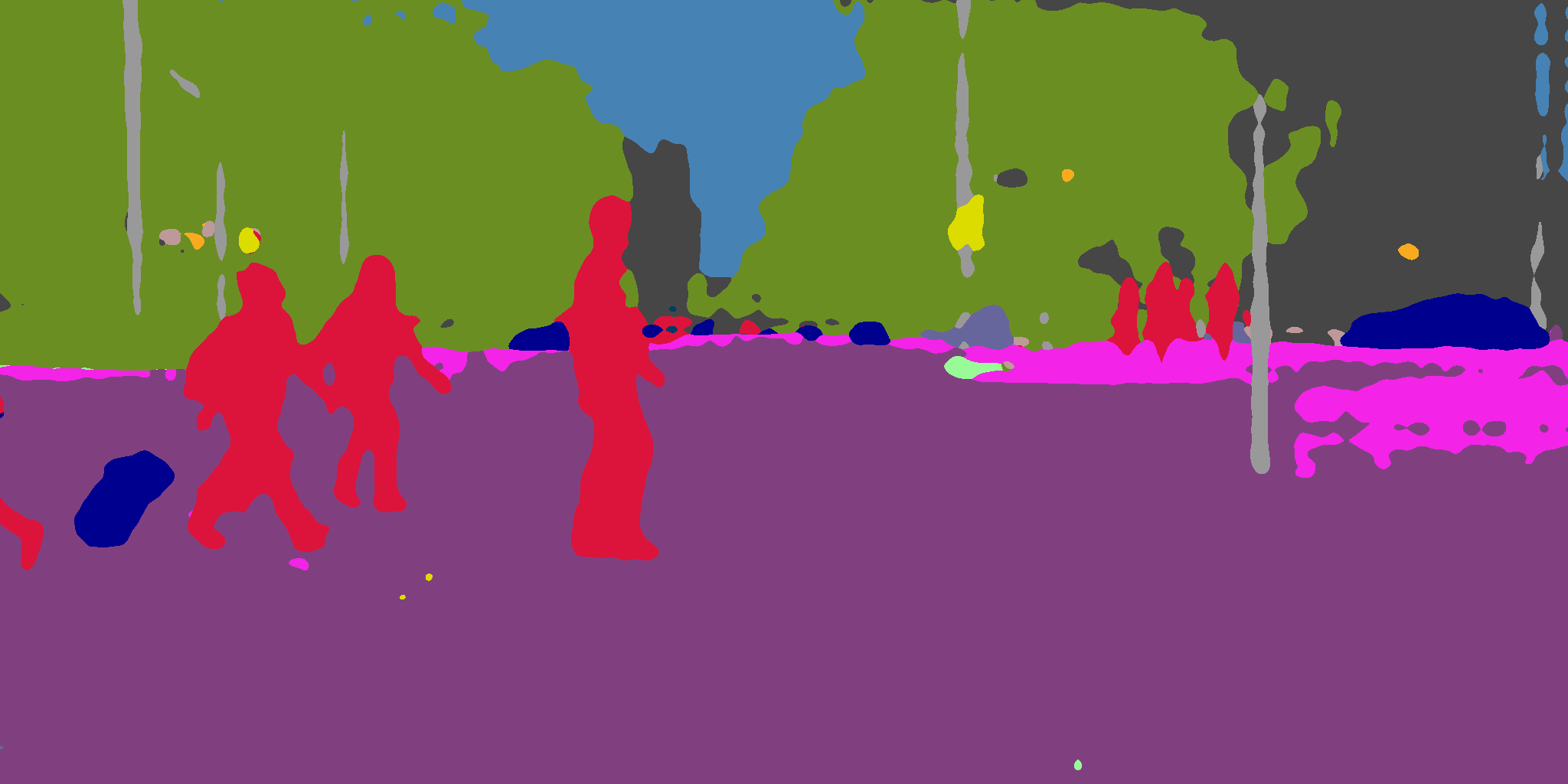}
    \\
    \includegraphics[width=0.22\textwidth]{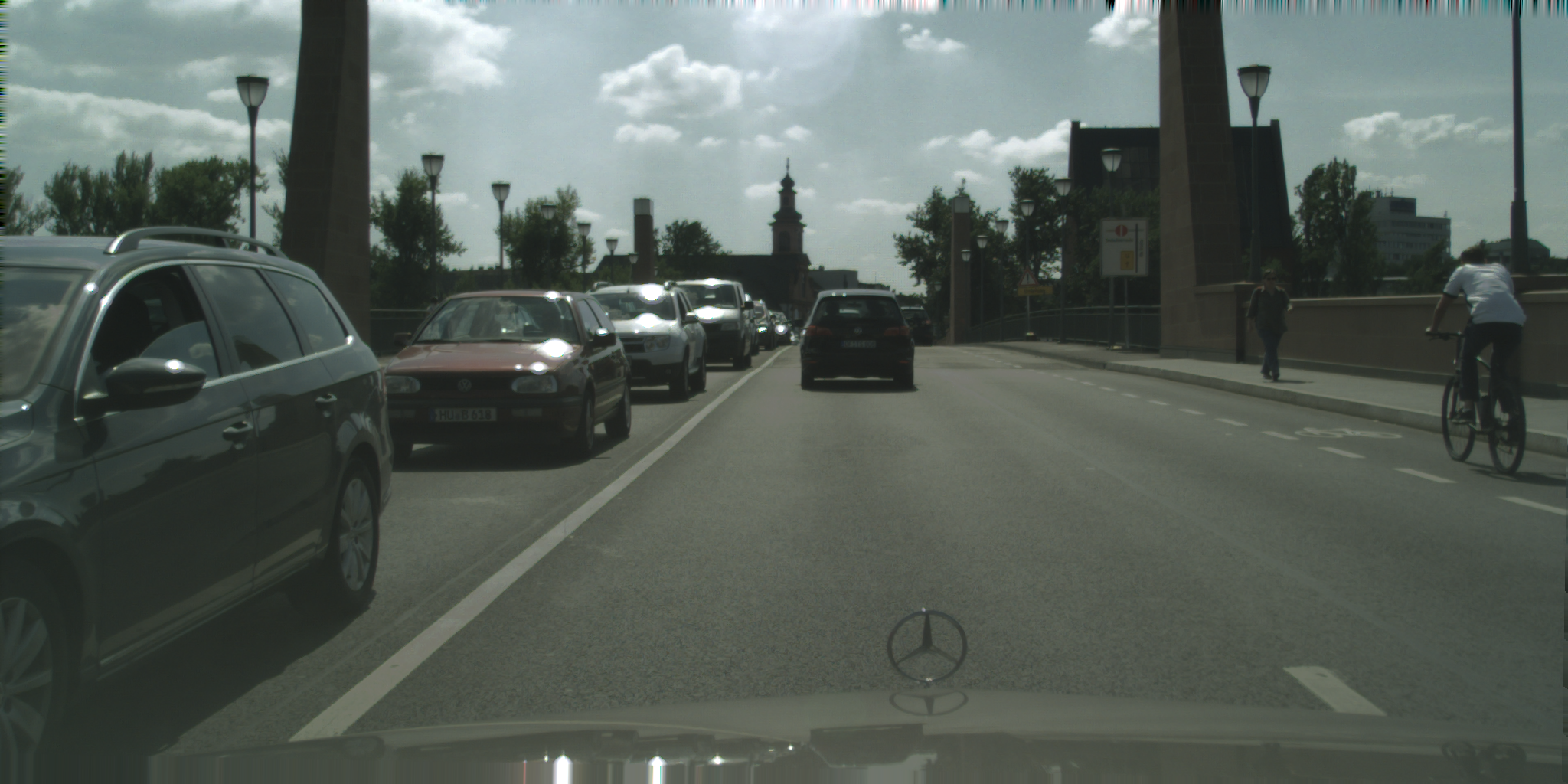} 
    &  
    \includegraphics[width=0.22\textwidth]{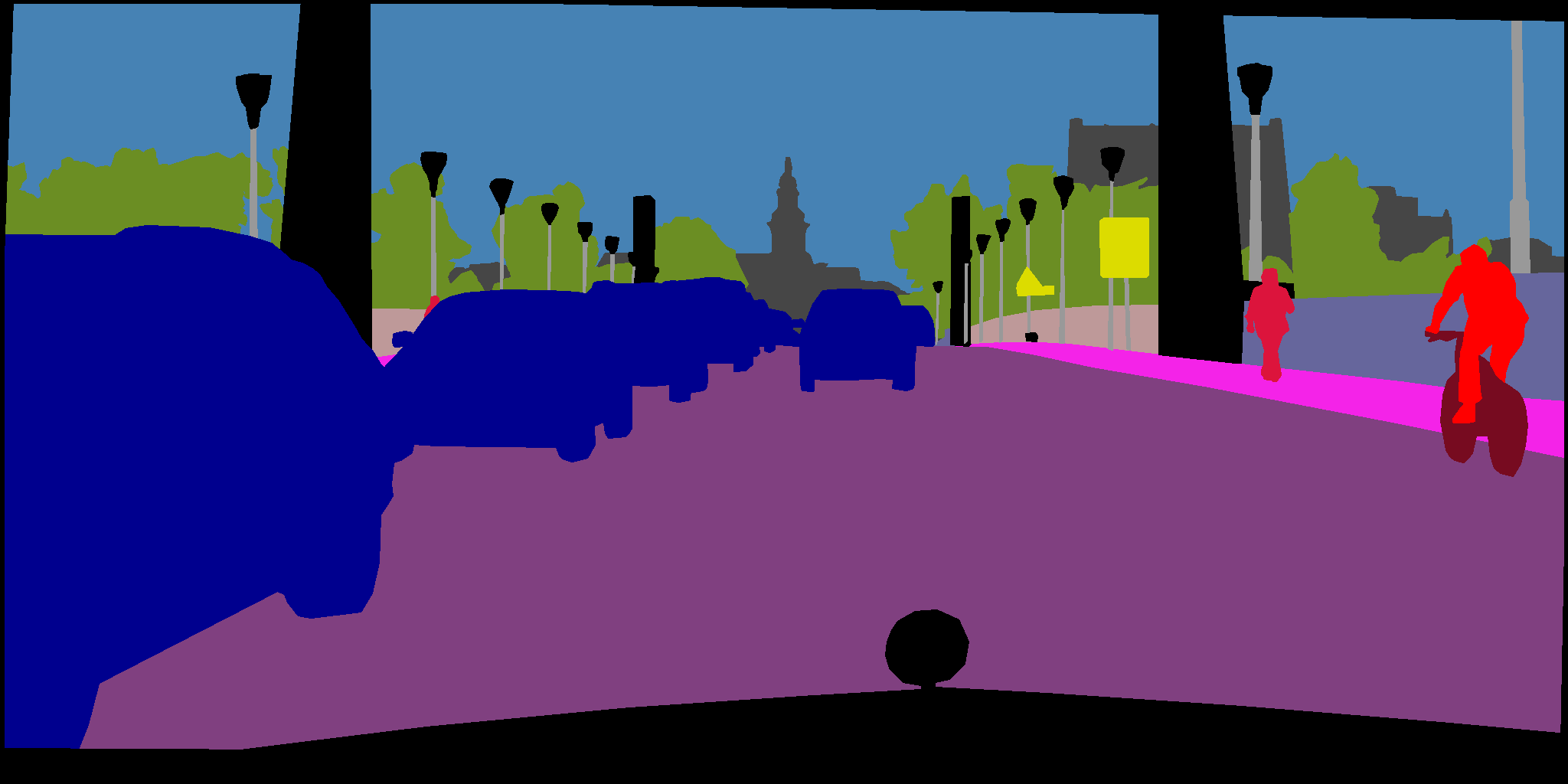}
    & 
    \includegraphics[width=0.22\textwidth]{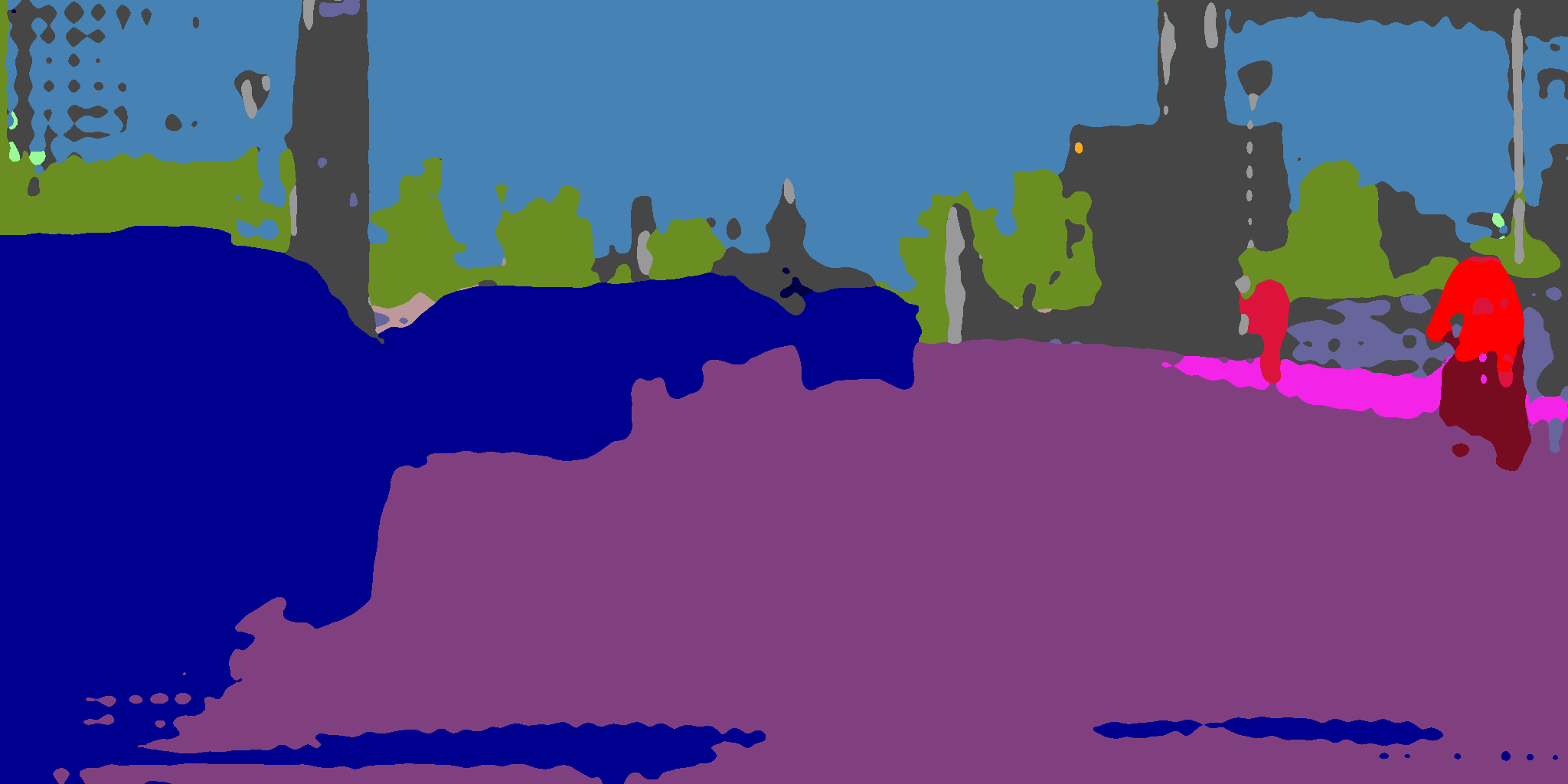}
    &
    \includegraphics[width=0.22\textwidth]{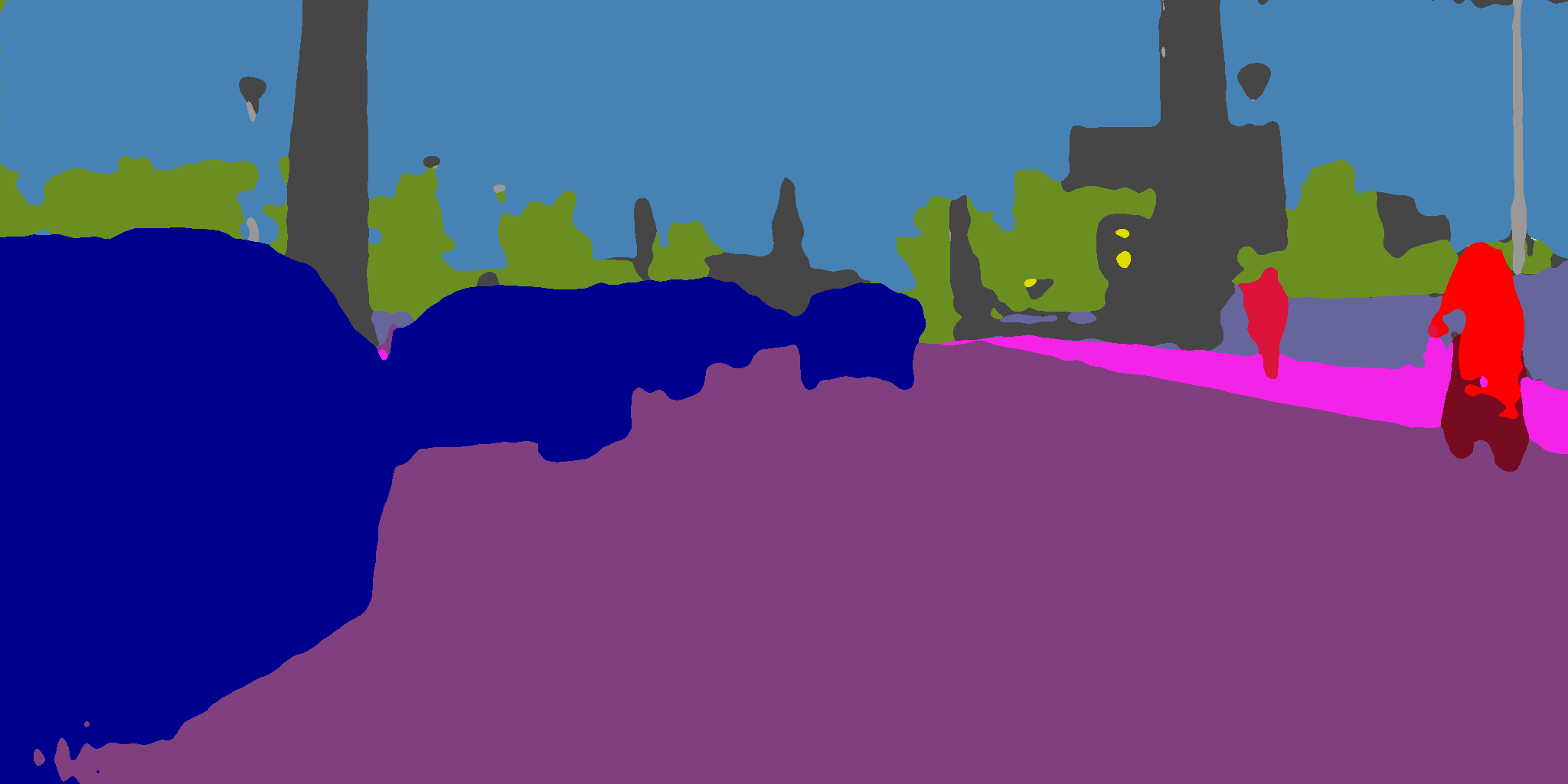}
    \\
    \centering Target
    & 
    \centering GT
    & 
    \centering SAT 
    & 
    \centering Ours
\end{tabular}
\caption{Visualization results on GTA5 to Cityscpaes. Black regions shown at the GT column are ignored during training.}
\label{fig: vis}
\end{figure*}
\section{Experiments}
\subsection{Overall Results}
We present overall adaptation results at Table \ref{tab:g2c} and Table \ref{tab:s2c} comparing with other UDA methods. We evaluate on two adaptation scenarios and the experiments show that our proposed methods is less complex but comparable with other UDA methods. For example, at Table \ref{tab:g2c}, \citet{pan2020unsupervised} consists of two pairs of segmentation network and discriminator compared with only one pair used by ours. \citet{Yang_2020_CVPR} utilizes three segmentation networks (MBT), two rounds self training (ST) and entropy regularization compared with only one segmentation network, one round self training and no entropy regularization used by ours. We also visualize the improvement of our methods at figure \ref{fig: vis}.
\begin{table}[htbp]
    \centering
    \small
    \begin{tabular}{clc}
    \toprule
    \multicolumn{3}{c}{GTA5$\to$Cityscapes}\\
    \midrule
     Backbone&Methods&mIoU\\
    \midrule
    \multirow{7}{*}{Resnet101}
    & \citet{li2018adaptive}&41.4\\
    & \citet{hoffman2018cycada}&42.7\\
    & \citet{zhang2020joint}&43.5\\
    & \citet{vu2019advent}&45.5\\\
    & \citet{toldo2021unsupervised}&45.9\\
    & \citet{pan2020unsupervised}&46.3\\
    & \citet{Yang_2020_CVPR} (MBT, T=0) & 46.8\\
    & \citet{Yang_2020_CVPR} (MBT, T=1) & 48.1\\
    & \citet{Yang_2020_CVPR} (MBT, T=2) & \textbf{50.4}\\
    &Ours&47.3\\
    \bottomrule
    \end{tabular}
    \caption{Overall results with comparison to other UDA methods on GTA5 to Cityscapes dataset. 19 classes are used to train and evaluate. (MBT, T=0): multiple networks with no self training; (MBT, T=1): multiple networks with one round self training; (MBT, T=2): multiple networks with two round self training}
    \label{tab:g2c}
\end{table}
\begin{table}[htbp]
    \centering
    \small
    \begin{tabular}{clc}
    \toprule
    \multicolumn{3}{c}{Synthia$\to$Cityscapes}\\
    \midrule
     Backbone&Methods&mIoU\\
    \midrule
    \multirow{7}{*}{Resnet101}
    &\citet{li2018adaptive}&46.7\\
    & \citet{hoffman2018cycada}&-\\
    &\citet{zhang2020joint}&48.3\\
    &\citet{vu2019advent}&48.0\\\
    &\citet{toldo2021unsupervised}&48.2\\
    &\citet{pan2020unsupervised}&48.9\\
    & \citet{Yang_2020_CVPR} (MBT, T=2) & \textbf{52.5}\\
    &Ours&48.9\\
    \bottomrule
    \end{tabular}
    \caption{Overall results with comparison to other UDA methods on Synthia to Cityscapes. 16 classes are used to train, and 13 classes are used to evaluated. (MBT, T=2): multiple networks with two round self training.}
    \label{tab:s2c}
\end{table}
\subsection{Ablation Study}
\begin{figure}[htp]
\centering
\includegraphics[width=0.7\columnwidth]{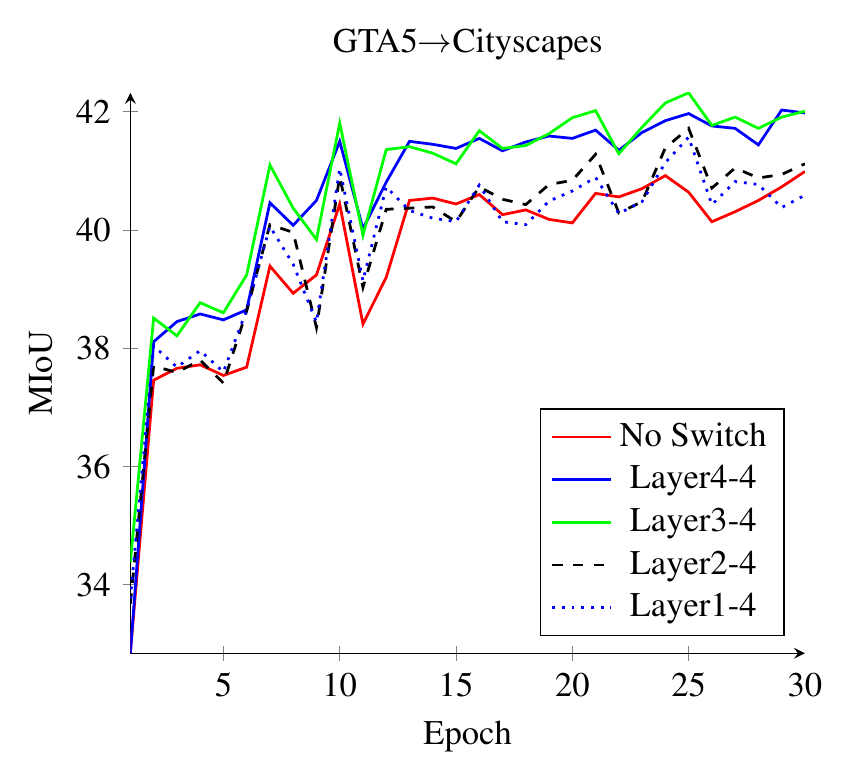}
\caption{Evaluation curve of model with No Switch compared with Switch (Layer i-j: switching from layer i to layer j). Experiments are done based on single level adaptation.}
\label{fig: abstudy}
\end{figure}

\textbf{Parameter Analysis:} Table \ref{tab: alpha} shows the sensitivity of $\alpha$. We could see that our proposed multi level adaptation ($\alpha\neq0$) is beneficial to domain alignment comparing with single level adaptation ($\alpha=0$).\\
\textbf{Layer Switch:} In normal situations, target Norm and AF are used only for inputs as no source data will be used during evaluation. However, we wonder how switching target Norm, AF with source ones in some layers affects model's performance. Thus, we do the following experiments. First, we divide the ResNet101 into 4 layers according to the place where downsampling happens. Second, we switch the target Norm and AF to source ones layer by layer and then do the evaluation. Due to limited computational resources, we only evaluate the switch starting from layer 4 to layer 1, which is shown at figure \ref{fig: abstudy}. What is beyond our expectation is that switching some certain layers even improves the model's performance, while there is 1.3 improvement at most on GTA5 to Cityscapes.
\begin{table}[bp]
    \centering
    \begin{tabular}{ccccccc}
    \toprule
    \multicolumn{7}{c}{GTA5$\to$Cityscapes}\\
    \midrule
    $\alpha$&0&0.01&0.05&0.1&0.2&0.4\\
    \midrule
    mIoU&41.0&41.9&\textbf{42.9}&42.5&42.6&41.7\\
    Range&0&+0.9&+1.9&+1.5&+1.6&+0.7\\
    \bottomrule
    \end{tabular}
    \caption{$\alpha$ Analysis. Experiments are done using Resnet101 as backbone with SEAT.}
    \label{tab: alpha}
\end{table}
\\
\textbf{Effectiveness:} In order to show the effectiveness and generally applicable of SEAT, we combine SEAT with other three adversarial learning based UDA methods and evaluate on GTA5 to Cityscapes adaptation scenario. The results are shown at Table \ref{tab: effective}. AdaStruct \cite{tsai2018learning} with single level adaptation, Advent (adv model) \cite{vu2019advent} , Differential \cite{wang2020differential} without self-training are improved to mIoU of 42.3, 44.0, 45.9 respectively.
\begin{table}[htp]
    \centering
    \begin{tabular}{lccc}
    \toprule
    \multicolumn{4}{c}{GTA5$\to$Cityscapes}\\
    \midrule
    Methods&mIoU&LS&Range\\
    \midrule
    \citet{tsai2018learning}$^*$&40.4&-&0\\
    +SEAT&41.0&-&+0.6\\
    +SEAT&\textbf{42.3}&$\checkmark$&+1.9\\
    \midrule
    \citet{vu2019advent}&43.8&-&0\\
    +SEAT&43.9&-&+0.1\\
    +SEAT&\textbf{44.0}&\checkmark&+0.2\\
    \midrule
    \citet{wang2020differential}$^+$&45.5&-&0\\
    +SEAT&44.1&-&-1.4\\
    +SEAT&\textbf{45.9}&\checkmark&+0.4\\
    \bottomrule
    \end{tabular}
    \caption{SEAT combined with other UDA methods on GTA5 to Cityscapes with ResNet101 as backbone. LS stands for layer switch. $^*$: reproduced result, 41.4 at the paper; $^+$: reproduced result, 46.2 at the paper.}
    \label{tab: effective}
\end{table}\\
\textbf{Modules Contributions:} Table \ref{tab:ab_g2s} and Table \ref{tab:ab_s2s} show the contribution of each module to the overall model performance for two adaptation scenarios.\\
For GTA5 to Cityscapes scenario, we firstly follow the work from \cite{tsai2018learning} with single level adaptation at the output space and achieve mIoU of 40.4, which is 1.0 lower than the paper shows. As mentioned at \cite{wang2020differential}, transferred target style source images is helpful to minimize the discrepancy of data distribution. Thus, we also adapt the transferred GTA5 images of \cite{hoffman2018cycada} to improve the mIoU to 43.4, which is served as the baseline for our works. Then, the mIoU is improved to 45.5 by adding the proposed SEAT (layer switch 3-4). After that, our proposed multi level adaptation is combined with $\alpha$ being set to 0.05 and the mIoU achieves 46.4 (layer switch 3-4). Finally, mIoU is improve to 47.3 with self training (layer switch 3-4).
\begin{table}[htp]
    \centering
    \begin{tabular}{lccccccc}
    \toprule
    \multicolumn{7}{c}{GTA5$\to$Cityscapes}\\
    \midrule
    Method&IT&SEAT&LS&MUL&ST&mIoU\\
    \midrule
    AA&&&&&&40.4\\
    +IT&\checkmark&&&&&43.3\\
    \midrule
    +SEAT&\checkmark&\checkmark&\checkmark&&&45.5\\
    +Mul&\checkmark&\checkmark&\checkmark&\checkmark&&46.4\\
    +ST&\checkmark&\checkmark&\checkmark&\checkmark&\checkmark&47.3\\
    \bottomrule
    \end{tabular}
    \caption{Ablation study on modules' contributions. AA represents Adversarial Adaptation; LS represents layer switch; IT stands for image style transferring; MUL represents multi level adaptation; ST represents self training.}
    \label{tab:ab_g2s}
\end{table}\\
As for Synthia to Cityscapes scenario, SEAT is added and the mIoU is increased to 47.0 (no layer switch) compared with mIoU of 44.6 from the baseline. After that, the mIoU is improved to 47.4 with our proposed multi level adaptaion ($\alpha=0.15$, layer switch 4-4). Finally, the mIoU is improved to 48.9 combined with self training (layer switch from 4-4).
\begin{table}[htp]
    \centering
    
    \begin{tabular}{lccccccc}
    \toprule
    \multicolumn{7}{c}{Synthia$\to$Cityscapes}\\
    \midrule
    Method&IT&SEAT&LS&MUL&ST&mIoU\\
    \midrule
    AA&&&&&&42.8\\
    +IT&\checkmark&&&&&44.6\\
    \midrule
    +SEAT&\checkmark&\checkmark&\checkmark&&&47.0\\
    +Mul&\checkmark&\checkmark&\checkmark&\checkmark&&47.4\\
    +ST&\checkmark&\checkmark&\checkmark&\checkmark&\checkmark&48.9\\
    \bottomrule
    \end{tabular}
    \caption{Ablation study on modules' contributions. AA represents Adversarial Adaptation; LS represents layer switch; IT stands for image style transferring; MUL represents multi level adaptation; ST represents self training.}
    \label{tab:ab_s2s}
\end{table}
\section{Conclusion}
In this paper, we propose Separate Affine Transformation and multi level adaptation, which is less complex compared with other UDA methods. Also, our proposed SEAT can be easily integrated to other existing adversarial learning based UDA methods and improve their model performance. By combining the SEAT and the proposed multi level adaptation with self training, we get comparable results with other UDA methods with less complexities.
\bibliography{main}
\end{document}